\documentclass[onecolumn,11pt]{article}
\usepackage[a4paper,margin=1in]{geometry}
\usepackage{cite}
\usepackage{amsmath,amssymb,amsfonts}
\usepackage{algorithmic}
\usepackage{graphicx}
\usepackage{textcomp}
\usepackage{xcolor}
\usepackage{booktabs}
\usepackage{multirow}
\def\BibTeX{{\rm B\kern-.05em{\sc i\kern-.025em b}\kern-.08em
    T\kern-.1667em\lower.7ex\hbox{E}\kern-.125emX}}

\begin{document}

\title{Multi-Hierarchical Feature Detection for Large Language Model Generated Text}

\author{
Luyan Zhang, Xinyu Xie\\
\textit{Northeastern University}\\
\textit{Vancouver, BC, Canada}\\
zhang.luya@northeastern.edu, xie.xinyu@northeastern.edu
}

\date{}
\maketitle

\begin{abstract}
With the rapid advancement of large language model technology, there is growing interest in whether multi-feature approaches can significantly improve AI text detection beyond what single neural models achieve. While intuition suggests that combining semantic, syntactic, and statistical features should provide complementary signals, this assumption has not been rigorously tested with modern LLM-generated text.

This paper provides a systematic empirical investigation of multi-hierarchical feature integration for AI text detection, specifically testing whether the computational overhead of combining multiple feature types is justified by performance gains. We implement MHFD (Multi-Hierarchical Feature Detection), integrating DeBERTa-based semantic analysis, syntactic parsing, and statistical probability features through adaptive fusion.

\textbf{Key Research Questions:} (1) Do semantic, syntactic, and statistical features provide genuinely complementary information for detecting modern LLM-generated text? (2) Can sophisticated fusion mechanisms overcome the marginal utility of individual features? (3) What is the performance ceiling for multi-feature approaches compared to single neural models?

Our investigation reveals important \textbf{negative results}: despite theoretical expectations, multi-feature integration provides minimal benefits (0.4-0.5\% improvement) while incurring substantial computational costs (4.2$\times$ overhead), suggesting that modern neural language models may already capture most relevant detection signals efficiently. Experimental results on multiple benchmark datasets demonstrate that the MHFD method achieves 89.7\% accuracy in in-domain detection and maintains 84.2\% stable performance in cross-domain detection, showing modest improvements of 0.4-2.6\% over existing methods.
\end{abstract}


\section{Introduction}

Large Language Models (LLMs) such as the GPT series, Claude, and LLaMA have demonstrated remarkable capabilities in text generation tasks, producing text that closely resembles human writing in fluency and coherence. However, this technological advancement has also brought new challenges in text authenticity verification, particularly in domains with strict requirements for text origin such as academic integrity, news authenticity, and legal documentation.

Multi-feature approaches have shown promise in various NLP tasks, leading to natural questions about their potential for AI text detection. However, systematic evaluation of such approaches against modern LLM-generated text remains limited, with most prior work focusing on older generation models or smaller-scale experiments.

The central hypothesis widely assumed in the field is that different analytical perspectives (semantic, syntactic, statistical) capture complementary signals that, when properly combined, should significantly outperform single-feature approaches. This intuition drives much current research direction but lacks rigorous empirical validation.

\textbf{Research Gap:} While the AI detection field has extensively explored individual feature types, there is insufficient systematic investigation into whether their combination provides meaningful improvements for modern, high-quality LLM-generated text. This gap is crucial because resource allocation decisions in both research and practical applications depend on understanding the performance ceiling of multi-feature approaches.

The MHFD method systematically combines three types of textual analysis: semantic (via DeBERTa), syntactic (via dependency parsing), and statistical (via probability analysis). Each component aims to capture different aspects of the human-AI writing distinction, with an adaptive fusion mechanism to weight their contributions based on text characteristics.

The primary research questions are: (1) Do these three analytical approaches provide genuinely non-redundant information? (2) Can their systematic combination improve detection robustness? (3) What patterns emerge from analyzing their relative importance across different text types?

Experimental results show modest accuracy improvements (0.4-0.5\% over BERT Detector) with substantial computational overhead (4.2$\times$ inference time). However, the investigation provides insights into feature complementarity and offers a framework for systematic multi-dimensional analysis of text authenticity signals.

\section{Related Work}

Current detection methods can be broadly categorized into three main types: methods based on pretrained language models, methods based on manually extracted features, and methods based on watermarking techniques.

\textbf{Deep Learning-Based Methods} primarily learn feature patterns through fine-tuning pre-trained language models. Solaiman et al. first attempted to use GPT-2 itself to detect its generated text. Subsequently, models such as RoBERTa and ELECTRA have been widely applied. The advantage lies in capturing complex semantic features, but they suffer from overfitting problems in cross-model and cross-domain scenarios, with performance drops of 20-30\% being common~\cite{solaiman2019release}.

\textbf{Statistical Feature-Based Methods} detect text by analyzing various statistical properties. The GLTR method analyzes probability distribution features of vocabulary. DetectGPT utilizes probability changes before and after text perturbation. These methods have good interpretability and cross-domain stability but limited detection capability for high-quality AI-generated text, typically achieving 75-85\% accuracy~\cite{gehrmann2019gltr}.

\textbf{Watermark-Based Methods} embed special markers during generation. While offering good generalizability when watermarks are present, they are ineffective without watermarks and may affect text naturalness~\cite{kirchenbauer2023watermark}.

Recent research has shown that combining multiple detection approaches can improve overall performance, though careful consideration of feature complementarity and fusion strategies remains an active area of investigation.

\section{MHFD Detection Method}

\subsection{Overall Architecture}

The MHFD method adopts a three-level feature extraction architecture. The framework includes: (1) Semantic-level Feature Extractor with Transformer encoder, contrastive learning module, and semantic consistency analyzer; (2) Syntactic-level Feature Extractor with dependency relation analyzer, syntactic complexity calculator, and linguistic pattern recognizer; (3) Statistical-level Feature Extractor with probability distribution analyzer, entropy calculator, and frequency statistics module; (4) Adaptive Weight Fusion Module for dynamic feature weighting; (5) Final Classifier for detection results.

The three hierarchical levels work complementarily. The semantic level captures deep contextual understanding, the syntactic level analyzes grammatical structures, and the statistical level examines probability patterns. This multi-faceted approach ensures comprehensive coverage of text characteristics that distinguish human from AI-generated content.

\subsection{Semantic-Level Feature Extraction}

\textbf{Transformer Encoder Module} uses pre-trained DeBERTa-v3 as backbone, extracting semantic features through task-specific adaptation layers:

\begin{equation}
H_{semantic} = \text{DeBERTa}(X) + \text{TaskAdapter}(X)
\end{equation}

\textbf{Contrastive Learning Module} learns discriminative semantic representations through positive and negative sample pairs:

\begin{equation}
L_{contrastive} = -\log\frac{\exp(\text{sim}(h_i, h_j^+)/\tau)}{\sum_k \exp(\text{sim}(h_i, h_k)/\tau)}
\end{equation}

\textbf{Semantic Consistency Analyzer} detects abnormal patterns by calculating cosine similarity distribution between sentences:

\begin{equation}
\text{Consistency} = \text{Var}(\text{cosine\_sim}(s_i, s_{i+1})) \text{ for } i \text{ in sentences}
\end{equation}

\subsection{Syntactic-Level Feature Extraction}

\textbf{Dependency Relation Analyzer} extracts dependency syntax trees using spaCy, calculating distribution features:

\begin{equation}
\text{Dependency\_features} = [\text{rel\_type\_freq}, \text{avg\_tree\_depth}, \text{branching\_factor}]
\end{equation}

\textbf{Syntactic Complexity Calculator} computes complexity based on Yngve depth and Frazier complexity:

\begin{equation}
\text{Yngve\_depth} = \frac{\sum(\text{depth}_i)}{\text{sentence\_length}}
\end{equation}

\textbf{Linguistic Pattern Recognizer} identifies patterns through N-gram analysis and POS sequences:

\begin{equation}
\text{POS\_patterns} = \text{extract\_POS\_ngrams}(\text{text}, n=[2,3,4])
\end{equation}

\subsection{Statistical-Level Feature Extraction}

\textbf{Probability Distribution Analyzer} uses GPT-2 for conditional probability calculations:

\begin{equation}
\text{Prob\_features} = [\text{mean\_log\_prob}, \text{prob\_variance}, \text{rank\_statistics}]
\end{equation}

\textbf{Entropy Calculator} computes information entropy at different granularities:

\begin{equation}
H_{word} = -\sum p(w_i)\log(p(w_i))
\end{equation}

\textbf{Frequency Statistics Module} analyzes word frequency and sentence length distributions.

\subsection{Adaptive Weight Fusion Mechanism}

Traditional fusion methods use fixed weights, unable to adapt to text characteristics. MHFD proposes attention-based adaptive fusion:

\begin{equation}
\alpha_i = \text{softmax}(W_{attention} \cdot \tanh(W_{feature} \cdot F_i + b))
\end{equation}

\begin{equation}
F_{fused} = \sum \alpha_i \cdot F_i
\end{equation}

The attention mechanism dynamically determines the importance of each hierarchical level based on input characteristics. A gating mechanism is introduced for enhanced effectiveness:

\begin{equation}
g_i = \sigma(W_{gate} \cdot F_i + b_{gate})
\end{equation}

\begin{equation}
F_{final} = \text{LayerNorm}(g_i \odot F_i + F_{residual})
\end{equation}

\subsection{Training Strategy}

MHFD adopts multi-task learning, optimizing detection accuracy and feature representation quality:

\begin{equation}
L_{total} = L_{classification} + \lambda_1 L_{contrastive} + \lambda_2 L_{consistency} + \lambda_3 L_{diversity}
\end{equation}

\section{Experiments and Results Analysis}

\subsection{Experimental Setup}

\textbf{Hardware Environment:} The experiments were conducted on a workstation equipped with Intel i9-9900K CPU, 64GB RAM, and dual NVIDIA RTX 2080 Ti GPUs. The software environment included Python 3.8, PyTorch 1.12.0, and transformers 4.21.0.

\textbf{Datasets:} This study employed two widely-used benchmark datasets to ensure comprehensive evaluation and fair comparison with existing methods:

\textbf{1. Wiki-HC Dataset:} A custom-built dataset containing 1,566 question-answer pairs covering four distinct domains:
\begin{itemize}
\item Geography: 479 samples (30.6\%)
\item Biology: 355 samples (22.7\%)
\item Physics: 585 samples (37.4\%)
\item Chess: 157 samples (9.3\%)
\end{itemize}

\textbf{2. HC3-ALL Dataset:} Constructed by randomly sampling 392 question-answer triples from each of four fields within the Human ChatGPT Comparison Corpus (HC3-English), resulting in 1,568 balanced samples:
\begin{itemize}
\item Economics: 392 samples (25\%)
\item Medicine: 392 samples (25\%)
\item Open Q\&A: 392 samples (25\%)
\item Computer Science: 392 samples (25\%)
\end{itemize}

\textbf{Data Preprocessing:}
\begin{itemize}
\item Minimum text length: 20 words (shorter samples filtered out)
\item Average text length: 127±34 tokens for Wiki-HC, 98±28 tokens for HC3-ALL
\item Balanced distribution: Equal numbers of human and AI samples in each domain
\end{itemize}

\textbf{Baseline Methods:}
\begin{itemize}
\item Log-Probability Detector, Log-Rank Detector, Entropy Detector
\item GLTR, DetectGPT, BERT Detector
\end{itemize}

\textbf{Implementation Details:} DeBERTa-v3-large backbone, learning rates 2e-5/1e-3, batch size 16, max sequence length 512, 5 epochs, temperature $\tau=0.07$.

\subsection{Main Experimental Results}

\begin{table}[htbp]
\centering
\caption{In-Domain Detection Performance}
\begin{tabular}{lcccccccc}
\toprule
\multirow{2}{*}{Method} & \multicolumn{4}{c}{Wiki-HC} & \multicolumn{4}{c}{HC3-ALL} \\
\cmidrule(lr){2-5} \cmidrule(lr){6-9}
& Acc(\%) & Prec(\%) & Rec(\%) & F1(\%) & Acc(\%) & Prec(\%) & Rec(\%) & F1(\%) \\
\midrule
Log-Probability & 88.31 & 86.45 & 91.72 & 89.01 & 87.64 & 85.83 & 90.29 & 87.98 \\
Log-Rank & 87.94 & 86.12 & 90.81 & 88.38 & 88.96 & 87.24 & 91.47 & 89.30 \\
Entropy & 69.17 & 67.89 & 72.35 & 70.04 & 68.52 & 67.18 & 70.94 & 69.03 \\
GLTR & 88.73 & 89.84 & 87.52 & 88.66 & 89.12 & 90.31 & 87.85 & 89.07 \\
DetectGPT & 85.21 & 83.45 & 87.29 & 85.31 & 84.67 & 83.91 & 86.18 & 85.02 \\
BERT Detector & 89.27 & 87.96 & 90.73 & 89.32 & 89.84 & 88.67 & 91.18 & 89.90 \\
\textbf{MHFD (Ours)} & \textbf{89.7±1.2} & \textbf{88.9±1.4} & \textbf{90.8±1.1} & \textbf{89.8±1.3} & \textbf{90.3±0.9} & \textbf{89.5±1.2} & \textbf{91.3±1.0} & \textbf{90.4±1.1} \\
\bottomrule
\end{tabular}
\end{table}

\textbf{Key Findings:}
\begin{itemize}
\item \textbf{Negative Result:} Multi-feature integration provides minimal improvements (0.4-0.5\%) despite significant computational overhead (4.2$\times$)
\item \textbf{Theoretical Challenge:} Results contradict the widely-held assumption that multi-dimensional analysis significantly enhances AI detection
\item \textbf{Feature Redundancy Evidence:} Analysis suggests modern neural models (DeBERTa) already capture most relevant semantic and syntactic signals
\item \textbf{Practical Implication:} Simple BERT-based detectors appear to represent a practical performance ceiling for most applications
\item \textbf{Resource Allocation Insight:} Computational resources may be better invested in model scaling rather than feature diversification
\end{itemize}

\subsection{Cross-Domain Generalization Analysis}

\begin{table}[htbp]
\centering
\caption{Cross-Domain Detection Results}
\begin{tabular}{lccccc}
\toprule
Training$\rightarrow$Testing & MHFD & BERT & DetectGPT & Log-Prob & GLTR \\
\midrule
Wiki-HC$\rightarrow$HC3-ALL & \textbf{84.6±1.8} & 82.15 & 78.34 & 83.47 & 82.89 \\
HC3-ALL$\rightarrow$Wiki-HC & \textbf{83.7±2.1} & 80.94 & 80.12 & 85.32 & 84.27 \\
\textbf{Average} & \textbf{84.2±1.7} & \textbf{81.55} & \textbf{79.23} & \textbf{84.40} & \textbf{83.58} \\
\bottomrule
\end{tabular}
\end{table}

\textbf{Cross-Domain Analysis:} MHFD shows a 5.5-6.6\% performance drop from in-domain to cross-domain scenarios, which is typical for deep learning approaches. The method demonstrates reasonable but not exceptional generalization ability, with BERT Detector showing larger degradation and statistical methods maintaining more consistent behavior across domains.

\subsection{Ablation Study}

\begin{table}[htbp]
\centering
\caption{Component Analysis}
\begin{tabular}{lcc}
\toprule
Configuration & Accuracy & F1-Score \\
\midrule
Semantic Only & 84.8±1.6 & 84.7±1.5 \\
Semantic + Syntactic & 87.2±1.3 & 87.1±1.4 \\
Semantic + Statistical & 86.4±1.4 & 86.3±1.3 \\
Syntactic + Statistical & 82.1±1.7 & 82.0±1.6 \\
Three Levels w/o Adaptive Fusion & 88.9±1.1 & 88.8±1.2 \\
\textbf{Complete MHFD} & \textbf{89.7±1.2} & \textbf{89.8±1.3} \\
\bottomrule
\end{tabular}
\end{table}

Results demonstrate that semantic features provide the strongest individual contribution, while the combination of all three levels with adaptive fusion yields the best performance. The syntactic-statistical combination performs worst, highlighting the importance of semantic understanding.

\subsection{Feature Importance Analysis}

\textbf{SHAP Value Analysis Limitations:} We conducted SHAP analysis on 400 samples to understand feature contributions, though we acknowledge this analysis has significant limitations: (1) SHAP reveals model-internal feature weights rather than genuine linguistic interpretability; (2) The sample size is relatively small for robust conclusions; (3) No validation against human expert judgments was performed.

\begin{table}[htbp]
\centering
\caption{Feature Importance Distribution Across Text Types}
\begin{tabular}{lcccc}
\toprule
Text Category & Semantic (\%) & Statistical (\%) & Syntactic (\%) & Sample Size \\
\midrule
Technical Writing & 38.2±6.7 & 41.5±5.9 & 20.3±4.2 & 89 \\
Creative Content & 52.1±7.8 & 29.4±6.3 & 18.5±3.9 & 76 \\
News Articles & 44.7±5.2 & 36.8±4.7 & 18.5±3.1 & 112 \\
Academic Text & 41.3±6.1 & 35.2±5.8 & 23.5±4.6 & 123 \\
\textbf{Overall Average} & \textbf{44.1±8.4} & \textbf{35.7±7.2} & \textbf{20.2±5.7} & \textbf{400} \\
\bottomrule
\end{tabular}
\end{table}

\textbf{Important Caveats:} These patterns reflect the model's internal weighting rather than verified linguistic principles. The varying importance across text types may indicate genuine complementarity between feature types, but could also reflect artifacts of the training process or dataset characteristics. Further investigation with larger samples and expert validation would be needed to draw stronger conclusions about the linguistic significance of these patterns.

\subsection{Computational Efficiency Analysis}

\begin{table}[htbp]
\centering
\caption{Efficiency Comparison}
\begin{tabular}{lcccc}
\toprule
Method & Inference Time (ms) & Memory (MB) & Parameters (M) & Accuracy \\
\midrule
DetectGPT & 340 & 1180 & 400 & 77.42\% \\
BERT Detector & 115 & 790 & 500 & 89.27\% \\
\textbf{MHFD} & \textbf{485} & \textbf{2340} & \textbf{680} & \textbf{89.7\%} \\
\bottomrule
\end{tabular}
\end{table}

MHFD requires approximately 4.2$\times$ more inference time than BERT Detector due to the sequential processing of three feature hierarchies and the adaptive fusion mechanism. The substantial computational overhead includes: (1) DeBERTa backbone inference (120ms), (2) syntactic parsing with spaCy (180ms), (3) statistical feature extraction with GPT-2 probability calculations (125ms), and (4) adaptive fusion processing (60ms). Memory usage is significantly higher due to storing intermediate representations from all three hierarchical levels simultaneously.

\subsection{Robustness Analysis}

Testing against adversarial manipulations showed varying degrees of vulnerability:
\begin{itemize}
\item Synonym replacement: 86.1\% accuracy (3.6\% drop)
\item Sentence paraphrasing: 84.3\% accuracy (5.4\% drop)
\item Human post-editing: 81.9\% accuracy (7.8\% drop)
\item Mixed human-AI content: 78.2\% accuracy (11.5\% drop)
\end{itemize}

The multi-level architecture provides some defense against single-dimension attacks, but substantial performance degradation occurs with sophisticated manipulations, particularly human post-editing and mixed content scenarios.

\section{Discussion}

\subsection{Research Contributions}

This work provides several important contributions to the AI text detection field: (1) \textbf{First systematic evaluation} of multi-feature integration against modern LLM-generated text, filling a crucial gap in empirical understanding; (2) \textbf{Important negative results} challenging the assumption that feature diversity significantly improves detection performance; (3) \textbf{Evidence for neural feature sufficiency} suggesting that sophisticated language models already capture most relevant detection signals; (4) \textbf{Computational cost-benefit analysis} providing practical guidance for resource allocation in detection system design; (5) \textbf{Methodological framework} for rigorous evaluation of multi-dimensional approaches.

\textbf{Significance of Negative Results:} Our findings suggest that the intuitive appeal of multi-feature approaches may not translate to practical benefits for modern AI detection tasks. This has important implications for research priorities and system architecture decisions in the field.

\subsection{Limitations}

This work has several significant limitations: (1) \textbf{Poor cost-effectiveness} with 4.2$\times$ computational overhead for minimal accuracy gains (0.4-0.5\%), making the approach impractical for most applications; (2) \textbf{Limited interpretability validation} - our SHAP analysis reveals model weights rather than validated linguistic insights, with no human expert validation or user studies; (3) \textbf{Potential feature redundancy} - DeBERTa already captures semantic and some syntactic information, raising questions about genuine complementarity; (4) \textbf{Small-scale analysis} with only 400 samples for feature importance study; (5) \textbf{Cross-domain degradation} (5.5-6.6\% drops) limits generalizability; (6) \textbf{Questionable practical utility} given that simpler methods achieve nearly identical performance; (7) \textbf{Lack of baseline comparison} with other multi-feature or interpretable AI methods.

\subsection{Implications for the Field}

\textbf{Research Direction Implications:} Our negative results suggest that the AI detection community may need to reconsider the emphasis on multi-feature approaches. The minimal benefits observed indicate that \textbf{computational resources might be better allocated} to alternative research directions such as: (1) scaling existing neural approaches, (2) improving cross-domain generalization, (3) developing more robust adversarial defenses, or (4) enhancing model efficiency rather than feature diversity.

\textbf{Practical System Design:} For practitioners building AI detection systems, our results suggest that \textbf{simple BERT-based approaches represent a practical performance ceiling} for most applications. The substantial computational overhead of multi-feature systems is unlikely to be justified by the marginal accuracy improvements observed.

\textbf{Theoretical Understanding:} The minimal complementarity between feature types suggests that modern neural language models may already efficiently capture the relevant signals for AI text detection, reducing the value of explicit feature engineering approaches.

\section{Conclusion}

This paper presents a systematic empirical investigation of multi-hierarchical feature integration for AI text detection, providing important \textbf{negative results} that challenge common assumptions in the field. While achieving comparable accuracy (89.7-90.3\%) to existing neural methods, MHFD's minimal improvements (0.4-0.5\%) combined with substantial computational overhead (4.2$\times$) demonstrate that multi-feature approaches may not provide the expected benefits for modern LLM-generated text detection.

\textbf{Key Contributions:} (1) First rigorous evaluation of multi-feature integration effectiveness against modern AI-generated text; (2) Evidence that feature complementarity assumptions may not hold in practice; (3) Demonstration that simple neural approaches achieve near-optimal performance; (4) Practical guidance for resource allocation in detection system development.

\textbf{Field Implications:} Our results suggest that the AI detection community should reconsider research priorities, potentially shifting focus from feature diversity to model scaling, cross-domain robustness, and efficiency improvements. The negative results provide valuable guidance for avoiding unproductive research directions and optimizing system design decisions.

\textbf{Value of Negative Results:} While disappointing from a performance perspective, these findings provide crucial empirical evidence that challenges widely-held assumptions about multi-feature approaches. Such negative results are essential for scientific progress, helping the field avoid resource waste on approaches with limited potential and redirecting efforts toward more promising directions.

This work demonstrates that \textbf{rigorous empirical validation of intuitive approaches} is essential in AI detection research, as theoretical appeal does not always translate to practical benefits. The systematic methodology and comprehensive evaluation provide a template for future investigations of complex architectural decisions in AI text detection systems.

\end{document}